\documentclass[10pt,conference,a4paper]{IEEEtran}

\usepackage{cite}
\usepackage{amsmath,amssymb,amsfonts}
\usepackage{algorithmic}
\usepackage{graphicx}
\usepackage{textcomp}
\usepackage{microtype}
\usepackage{booktabs}
\usepackage{tabularx}
\usepackage{xcolor}
\def\BibTeX{{\rm B\kern-.05em{\sc i\kern-.025em b}\kern-.08em
    T\kern-.1667em\lower.7ex\hbox{E}\kern-.125emX}}
\begin{document} 

\title{How important are faces for person re-identification?}


\author{\IEEEauthorblockN{Julia Dietlmeier, Joseph Antony, Kevin McGuinness, Noel E. O’Connor}
\IEEEauthorblockA{Insight SFI Research Centre for Data Analytics, Dublin City University\\\small
\{julia.dietlmeier, joseph.antony, kevin.mcguinness, noel.oconnor\}@insight-centre.org}

}

\maketitle

\begin{abstract}
This paper investigates the dependence of existing state-of-the-art person re-identification models on the presence and visibility of human faces. We apply a face detection and blurring algorithm to create anonymized versions of several popular person re-identification datasets including Market1501, DukeMTMC-reID, CUHK03, Viper, and Airport. Using a cross-section of existing state-of-the-art models that range in accuracy and computational efficiency, we evaluate the effect of this anonymization on re-identification performance using standard metrics. Perhaps surprisingly, the effect on mAP is very small, and accuracy is recovered by simply training on the anonymized versions of the data rather than the original data. These findings are consistent across multiple models and datasets. These results indicate that datasets can be safely anonymized by blurring faces without significantly impacting the performance of person re-identification systems, and may allow for the release of new richer re-identification datasets where previously there were privacy or data protection concerns.
\end{abstract}


\section{Introduction}

The task of person re-identification (re-ID) is to retrieve images of a specified individual in a large non-overlapping multi-camera database, also called gallery, given a query person of interest. It is an important component in intelligent video surveillance systems, which can be used to improve public safety with the increasing number of surveillance cameras in University campuses, theme parks, streets, train stations, and airports~\cite{6_Ye_ReID_Survey_arXiv2020}. In addition, tracking with person \text{re-ID} can provide useful information for improving city planning e.g. for retail space management. Current advances in re-ID can be attributed to deep learning and the availability of large public image- and video-based re-ID datasets. Deep learning, in particular, has significantly improved the accuracy of re-ID systems, but several recent studies suggest that the number and scale of available datasets is insufficient~\cite{8_Grigorev_UAVimages_2019,9_Zheng_ReID_arXiv2018}.

\begin{figure*}[t]
\centering
\includegraphics[scale=0.6]{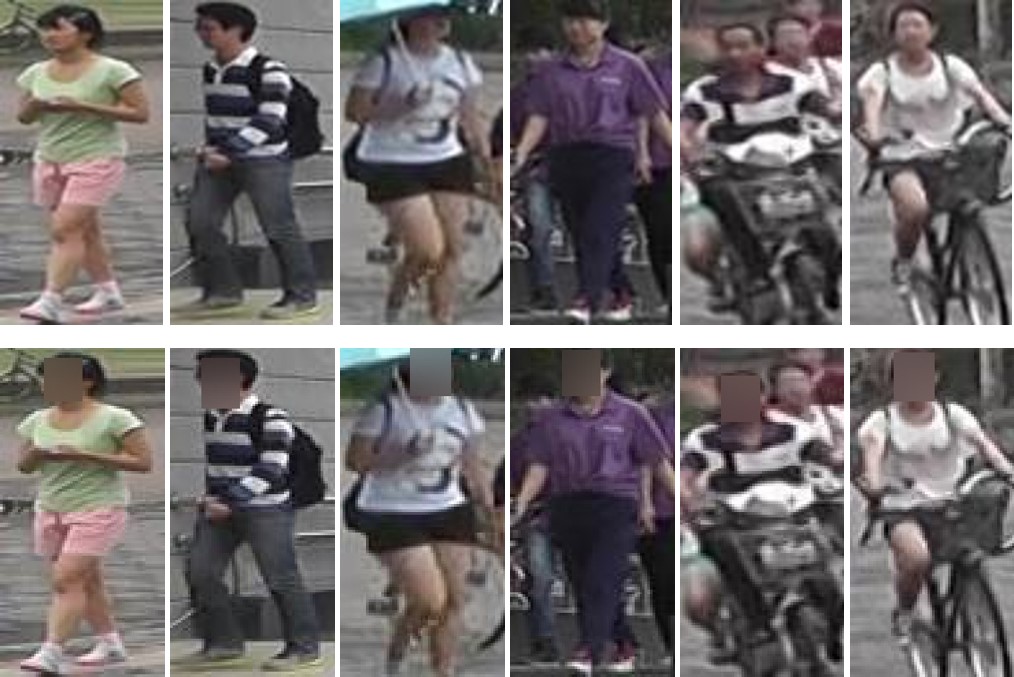}
\caption{Examples of original (first row) and our anonymized (second row) images from the Market1501 re-ID dataset. For face de-identification, we apply the TinyFaces~\cite{Tiny_Faces_CVPR2017} face detector first and then blur the detected region with a large kernel Gaussian to remove all privacy-sensitive information. }
\label{fig:anon_intro}
\end{figure*}

The release of new person re-identification datasets can raise legitimate privacy concerns. As an example, 97\% of the people surveyed by Harvard Business Review~\cite{3_Morey_Harvard_2015} expressed concern that businesses and the government might use their data inappropriately. Privacy is one of the most important political and social concepts in our society and it is vital to keep sensitive data out of the wrong hands where it may be used for personal stalking, harassment, identity theft, blackmail, and mass surveillance~\cite{1_Rocher_NatureComms_2019,7_Heshmaty_DataMisuse_2019}. In relation to the data misuse, article 5(1)(b) of European General Data Protection Regulation (GDPR) states: ``Personal data shall be collected for specified, explicit and legitimate purposes and not further processed in a manner that is incompatible with those purposes.''
 
Sensitive data and its use is protected under law. One way to comply with the strict data privacy regulations such as the California Consumer Privacy Act (CCPA) and the GDPR is data anonymization. Effective anonymization can allow data to be released to benefit society without compromising the identities of the individuals that appear in the data. Data protection laws such as GDPR do not apply to personal data that have been anonymized~\cite{1_Rocher_NatureComms_2019,2_EU_Regulation_2016}. This fact assures that data breaches with associated fines can be avoided and that data can be used, shared, and sold. 

The main biometric personal identifier in images and videos is the human face. Therefore, in the context of person \text{re-ID}, the anonymization process of a dataset could involve de-identification of individual faces. De-identification is common in medical imaging~\cite{deidentify_medical} and relates to the removal of identifying information from images prior to sharing of  data. Such information could, for example, potentially be used to derive a person's identity. There are several studies on image or video-based face de-identification~\cite{4_Ren_FaceAnonPrivacy_arXiv2018,5_Hukkelas_DeepPrivacy_2019,10_Gross_Face_CVPR2008,12_Sweeney_kanonymity_2002,13_Newton_Faces_IEEE2005,14_Gafni_FaceAnonVideos_ICCV2019,16_Sun_Obfuscation_CVPR2018,17_Sun_Obfuscation_ECCV2018,18_Gross_CVPRW_2006,19_Gross_bookSpringer_2009}. While person re-ID itself is a well-developed field and significant progress has been achieved with modern deep learning-based methods~\cite{6_Ye_ReID_Survey_arXiv2020,20_ReID_Survey_Neurocomputing_2019,21_Lavi_ReID_survey_arXiv2018}, privacy-preserving face de-identification has received comparatively little attention~\cite{5_Hukkelas_DeepPrivacy_2019}. 

Two years on from the creation of European GDPR, the issue of data privacy contributes even more to the quest for a detailed investigation into the impact of anonymization on the performance of modern person re-ID algorithms. It is important to understand to which extent the performance of person re-ID algorithms is diminished by anonymization. 

This research investigates for the first time the impact of anonymization on the accuracy of person re-ID systems. We benchmark the performance of various state-of-the-art person \text{re-ID} models on five public datasets: Market1501~\cite{Market1501_ICCV_2015}, DukeMTMC-reID~\cite{DukeMTMC_CVPRW_2017}, CUHK03 \cite{CUHK03_CVPR_2014}, VIPeR~\cite{VIPeR_ECCV_2008} and Airport \cite{Airport_IEEE_2017}, and quantify the performance penalty of the face anonymization procedure. Our results show that face anonymization in benchmark image-based datasets has only a marginal effect on model performance. This finding could be important in the decision to release new anonymized re-ID datasets without breaching privacy regulations or hindering technology advancement. 


\section{Related Work}
\textbf{Face anonymization.} Face anonymization or de-identification aims at removing privacy-sensitive information from detected faces. Existing approaches to face de-identification can be categorized into naïve, the k-same family of algorithms, and the methods based on Generative Adversarial Networks (GANs). 

Naïve methods apply image processing techniques such as pixelation (image/face subsampling), black-out or blurring (smoothing the image or the detected face with e.g. a Gaussian filter with large kernel). There are concerns being raised that these methods fail in removing privacy-sensitive information in some cases and no formal privacy guarantees can be made~\cite{13_Newton_Faces_IEEE2005,18_Gross_CVPRW_2006,5_Hukkelas_DeepPrivacy_2019, 19_Gross_bookSpringer_2009}. Gross et al.~\cite{19_Gross_bookSpringer_2009} e.g. show that blurring and pixelation can successfully outsmart human recognition. But these methods, according to the authors, lack a privacy model and are susceptible to relatively simple attacks. An example of a naïve method (blur) is shown in Figure~\ref{fig:anon_intro}. We provide a discussion on the blurring technique and its resistance to deblurring and other attacks at the end of this paper.


\begin{table*}[t]
\caption{\label{tab1}Reported performance of selected state-of-the-art person re-ID models.}
\centering\small
\begin{tabular}{l l c c c c}
\toprule
Models & Venue & \multicolumn{2}{c}{Market1501} & \multicolumn{2}{c}{DukeMTMC-reID}\\
\cmidrule{3-6}
& & mAP & Rank1 & mAP & Rank1 \\
\midrule
BoT \cite{Luo_BoT_model_CVPRW_2019}          & CVPRW 2019 & 94.2 & 95.4 & 89.1 & 90.3\\ 
PCB \cite{Sun_PCB_ECCV_2018}                 & ECCV 2018  & 81.6 & 93.8 & 69.2 & 83.3 \\
MLFN \cite{Chang_MLFN_CVPR_2018}             & CVPR 2018  & 74.3 & 90.0 & 62.8 & 81.0 \\ 
HACNN \cite{Li_HACNN_CVPR_2018}              & CVPR 2018  & 75.7 & 91.2 & 63.8 & 80.5 \\ 

Resnet50Mid \cite{Yu_Resnet50Mid_arXiv_2017} & arXiv 2017 & 75.6 & 89.9 & 63.9 & 80.4 \\ 
\bottomrule
\end{tabular}
\end{table*}
 
The k-anonymity algorithm of Sweeney~\cite{12_Sweeney_kanonymity_2002} lays the theore-tical foundation for the category of k-same de-identification of face images. Newton et al.~\cite{13_Newton_Faces_IEEE2005} are the first to develop the privacy-enabling k-same algorithm that preserves many facial details and at the same time guarantees that face recognition software cannot recognize de-identified faces. The algorithm computes similarity between faces based on a distance metric. Further, it creates new faces by averaging image components, which may be the eigenvectors or the original image pixels. Gross~et~al.~\cite{10_Gross_Face_CVPR2008} design a re-coding system that separates out identity and non-identity related components based on semi-supervised learning of multi-factor models.

With the advent of GANs~\cite{15_Goodfellow_GANs_NIPS2014} it is possible to generate \text{images} that often resemble the real data distribution. Hukkelas et al.~\cite{5_Hukkelas_DeepPrivacy_2019} show that GANs are an efficient tool to remove all privacy-sensitive information without destroying the original image quality. Other recent works on GAN-based methods~\cite{16_Sun_Obfuscation_CVPR2018, 17_Sun_Obfuscation_ECCV2018} consider the related task of person obfuscation. Sun et al.~\cite{16_Sun_Obfuscation_CVPR2018} propose a GAN-based method to complete head regions based on the context and their obfuscation method is designed to work against any human or machine recognizer. The method in~\cite{17_Sun_Obfuscation_ECCV2018} uses a combination of a parametric face model reconstruction and GAN-based image synthesis. This method provides control over facial parameters for manipulation of identity and enables photo-realistic image synthesis. The recent architecture of Gafni et al.~\cite{14_Gafni_FaceAnonVideos_ICCV2019} is based on an adversarial autoencoder and a trained face classifier. The key objective is to decorrelate the identity and to fix the expression, pose, and illumination components. This method for face de-identification allows automatic video modification at high frame rates. 

To our best knowledge, there have not been any prior studies on face anonymization in the context of person re-ID.


\textbf{Image-based person re-ID.}
Approaches to person re-identification can be categorized into traditional and deep learning-based methods.  Traditional methods aim to design hand-crafted features and learn an effective distance metric. Convolutional neural networks (CNNs), recurrent neural networks (RNNs), and GANs form the basis for deep learning-based person \text{re-ID}. Wu et al.~\cite{20_ReID_Survey_Neurocomputing_2019} highlight the fact that almost all methods for person re-identification in three top conferences (CVPR, ICCV, ECCV) in recent years are based on deep learning. The most recent survey by Ye at al.~\cite{6_Ye_ReID_Survey_arXiv2020} provides comprehensive analysis of existing deep learning methods by analyzing their advantages and drawbacks. Here, we only review the recent contributions in deep learning-based re-ID technology that are used in this paper. Table~\ref{tab1} summarizes the performance of these approaches on two common benchmark datasets such as Market1501 and DukeMTMC-reID.

We start with the bag of tricks model (BoT) \cite{Luo_BoT_model_CVPRW_2019} 
which leverages a number of effective training ``tricks'' in person re-ID. The model architecture can remain unchanged as most tricks can be added to the standard baseline. These training tricks include:
\begin{itemize}
    \item a warmup learning rate;
    \item random erasing augmentation (REA) as initially proposed by Zhong et al.~\cite{Zhong_REA_AAAI_2017};
    \item label smoothing (LS)~\cite{Szegedy_LS_CVPR_2016}, a method used to prevent overfitting;
    \item changing the last stride (last spatial down-sampling operation in the backbone network) from 2 to 1 results to provide higher spatial feature
resolution;
    \item addition of a batch normalization (BN) layer between feature extraction and fully connected classifier layers; and,
    \item use of the center loss~\cite{Wen_CenterLoss_ECCV_2016}.
\end{itemize}
Luo et al.~\cite{Luo_BoT_model_CVPRW_2019} use these tricks to surpass human-level performance using only global features without the part constraints in the BoT model and report 95.4\% rank-1 accuracy and 94.2\% mAP on Market1501 (see Table~\ref{tab1}). 

Sun et al.~\cite{Sun_PCB_ECCV_2018} address the problem of learning discriminative part-informed features for person retrieval with the development of their PCB model. This model produces a convolutional descriptor consisting of several part-level features. Instead of partitioning the input image, the PCB partitions the 3D tensor $\mathbf{T}$ of activations into $p$ pieces of column vectors $\mathbf{g}$. PCB is essentially a classification network with small modifications on the backbone network.

Yu et al.~\cite{Yu_Resnet50Mid_arXiv_2017} (Resnet50Mid model) leverage mid-level features from the earlier layers of a residual deep neural network. The authors claim that learning discriminative view-invariant features from multiple semantic levels is very important in deep learning-based person \text{re-ID}. Color and texture are low-level semantic concepts. High-level semantic concepts can be shape and gender, for example. The Resnet50Mid model combines the mid-layer feature maps with the final-layer feature map.

The idea of using multiple semantic levels is also followed by Chang et al.~in~\cite{Chang_MLFN_CVPR_2018}. The MLFN model (Multi Level Factorization Net) developed can learn identity-discriminative and view-invariant visual factors at multiple semantic levels. The MLFN architecture is tailored to identify discriminative latent factors in input images. Different levels of the network generate latent attributes of different semantic levels.

The harmonious attention convolutional neural network (HACNN)~\cite{Li_HACNN_CVPR_2018} is a lightweight architecture developed by using a combination of a novel attention mechanism and feature representation. This network can learn hard region-level together with soft pixel-level attention inside person bounding boxes. The correlated complementary information between attention selection and feature discrimination is maximized by using re-ID feature representations.


\begin{table*}[t]
\caption{\label{tab_datasets}Statistics of the image-based person re-ID datasets used in the experimental part of this paper.}
\centering \small
\begin{tabular}{l c c r r r r}
\toprule
Dataset & Released in & Cameras & Identities & Training images & Gallery images & Query images \\
\midrule
Market1501 \cite{Market1501_ICCV_2015}                      & 2015 & 6 & 1,501 & 12,936 & 19,732 & 3,368 \\ 

DukeMTMC-reID \cite{DukeMTMC_CVPRW_2017}                    & 2017 & 8 & 1,404 & 16,522 & 17,661 & 2,228\\

CUHK03 (detected) \cite{CUHK03_CVPR_2014}                              & 2014 & 6 & 1,360 & 7,365 & 1,400 & 5,332 \\ 

Airport \cite{Airport_IEEE_2017},\cite{Airport_TPAMI_2018}  & 2018 & 6 & 770 & 3,493 & 2,420  & 1,003 \\ 

VIPeR  \cite{VIPeR_ECCV_2008}                               & 2008 & 2 & 632 &  316 & 316 & 316 \\ 
\bottomrule

\end{tabular}
\end{table*}
Despite their merits, the above person re-identification models match people only at a single scale. To alleviate this problem, Qian et al.~\cite{Qian_MuDeep_ICCV_2017} propose a novel multi-scale deep learning model (MuDeep). Multi-scale learning is adopted by the MuDeep model to learn discriminative features at different spatial scales and locations. Unlike all above models, the authors of MuDeep do not evaluate on Market1501 and DukeMTMC-reID datasets.

There are many more state-of-the-art person re-ID models which are covered in some recent reviews~\cite{6_Ye_ReID_Survey_arXiv2020,20_ReID_Survey_Neurocomputing_2019}. 

\section{Methodology}

\subsection{Face detection} Face detection in person re-ID datasets must deal with significant challenges associated with large variances in poses of the face, non-uniform illumination and most of all small face sizes in low resolution images. Based on this observation, we decide to use a PyTorch implementation~\cite{Pytorch_tiny_faces} of the TinyFaces detector~\cite{Tiny_Faces_CVPR2017}. The tiny face detector can work with very small faces and reduces  error by a factor of two compared with prior methods on the WIDER FACE dataset. However, even TinyFaces is not 100\% accurate. As an example, consider the Airport dataset. The query set of the Airport dataset contains 1003 images and the number of missed detections amounts to around 100 images. Inspection of 500 of the 2420 gallery images shows the detector misses approx.~40 faces, indicating approximately 90\% recall in this dataset. 

The CUHK03 dataset contains 28k images. Visual inspection of the first 1000 reveals that the detector misses less than 20 faces. Assuming the first 1000 images represent an unbiased sample, then we have approx.~98\% recall. Using the same methodology we find that face detection has approx.~90\% recall on Market1501 and DukeMTMC-reID datasets. 

The VIPeR dataset is immensely challenging due to small data size and low image resolution~\cite{Qian_MuDeep_ICCV_2017}. The dataset consists of two folders \text{cam\_a} and \text{cam\_b} and the training, query, and gallery splits are defined in the supplied JSON file. TinyFaces does not detect many faces in the VIPeR dataset with the default probability threshold (only 5.7\% of images in \text{cam\_a} have a face detected with a threshold of 0.65). This is likely due to the low resolution of the images and the fact that many images  have side-profile faces or no face (back of the head). Lowering the threshold to 0.05, TinyFaces detects faces in 41.77\% of images in \text{cam\_a} and 8.23\% on \text{cam\_b} (\text{cam\_b} contains many more images with side and back profiles). Since false positives do not pose a problem, in cases with no detections we add a default detection with bounding box $x_0, y_0, x_1, y_1 = (15, 4, 35, 23)$. This default detection is selected based on the average detected face location in \text{cam\_a}, adding 4 pixels to the width of the box to account for the larger variance in the $x$-direction.

\subsection{Face anonymization} Face de-identification in each dataset is performed by blurring. We use a pre-trained state-of-the-art TinyFaces model to detect faces and locate a bounding box for each face. A crop of the bounding box is then extracted and blurred using a Gaussian blur filter with radius equal to $1/8$ the width of the size of the bounding box. This method is chosen to ensure that the filter radius scales with the size of the bounding box and ensure that a stronger blur is applied to higher resolution face patches. Experimentally $1/8$ is verified to effectively remove all identifying face features. Figure~\ref{fig:anon_intro} illustrates the results of this procedure.

\subsection{Datasets}
We evaluate the effect of anonymization on five popular person re-identification datasets: Market1501~\cite{Market1501_ICCV_2015}, DukeMTMC-reID~\cite{DukeMTMC_CVPRW_2017}, CUHK03~\cite{CUHK03_CVPR_2014}, VIPeR~\cite{VIPeR_ECCV_2008}, and Airport~\cite{Airport_TPAMI_2018}. Table~\ref{tab_datasets} summarizes statistics for these datasets.

\textbf{Market1501} contains 32,668 images of 1,501 pedestrians, each of which is recorded by at most six cameras placed in front of a campus supermarket. There are $3.6$ images on average at each viewpoint for each person. Background clutter and the misalignment problem are some of the re-ID challenges due to all dataset images being cropped using a deformable part model (DPM) detector.

\textbf{DukeMTMC-reID} is a relatively new, manually annotated, calibrated, multi-camera data set recorded outdoors on the Duke University campus. It was captured by eight non-overlapping camera views. It contains 1,404 identities captured in more than two cameras along with 408 identities (distractors) in only one camera, 16,522 training images, 17,661 gallery images, and 2,228 queries.

\textbf{CUHK03} consists of 13,164 images of 1,360 pedestrians recorded by six cameras. Each identity is captured from two disjoint camera views. There are an average of 4.8 images in each view. DPM-detected and hand-labeled bounding boxes are both provided. Misalignment, occlusions, and missing body-parts are common re-ID challenges in this dataset as it is close to a realistic setting. 

\textbf{VIPeR}, despite being one of the earlier released dataset, is considered one of the most challenging. It consists of 632 pedestrian image pairs taken from arbitrary viewpoints under varying illumination conditions. This dataset contains two cameras, each of which captures one image per person. It also provides the viewpoint angle of each image. Each image is resized to $128\times 48$ pixels.

\textbf{Airport} was captured from six cameras of an indoor surveillance network in a mid-sized airport in Cleveland, USA~\cite{Airport_IEEE_2017,Airport_TPAMI_2018}. Each camera has $768\times 432$ pixels and captures video at 30 frames per second. 12-hour long videos from 8 AM to 8 PM were collected and each video was randomly spit into 40 five minute long clips. This dataset captures time-varying crowd dynamics and thus differs from the other datasets in the temporal aspect. It is useful to evaluate the temporal performance of the re-ID algorithms. There are 9,651 identities, with 39,902 bounding boxes in total with an average of 3.13 images per person.

\section{Experiments}
\subsection{Implementation details}
We use the Pytorch-based Torchreid library developed by Zhou and Xiang~\cite{torchreid} for the PCB, MLFN, Resnet50Mid, HACNN, and MuDeep models. We implement the BoT model according to~\cite{BoT_implementation}.

\textbf{Training.} We train the PCB, MLFN, Resnet50Mid, HACNN, and MuDeep models for 100 epochs using the cross entropy loss. Horizontal random flip and random 2D translation augmentation methods are used during training where the image size is first increased to $(1+ \frac{1}{8})$ with the probability of $p=0.5$ and then the random crop is performed. The number of parts in the PCB model is set to six. Adam is used as the optimizer with default hyperparameters of $\beta_1=0.9$ and $\beta_2=0.999$. The initial learning rate is set to $0.003$ and the learning rate decay is $\gamma=0.1$. The batch size is 32 by default.

We train the BoT model for 120 epochs using both cross entropy loss and a triplet loss. We use a Resnet50 with pre-trained on ImageNet as the backbone network. Adam is used to optimize the model. The warmup learning rate strategy for the BoT model is as follows: first, the learning rate linearly increases from $3.5 \times 10^{-5}$ to $3.5 \times 10^{-4}$ in the first 10 epochs. Then, the learning rate is decayed to $3.5 \times 10^{-5}$ and $3.5 \times 10^{-6}$ at 40th epoch and 70th epoch respectively. The learning rate $lr(t)$ at epoch $t$ is computed as:
\[
 lr(t) = 
  \begin{cases} 
   3.5 \times 10^{-5} \times \frac{t}{10} & \text{if } t \leq 10 \\
   3.5 \times 10^{-4}                     & \text{if } 10 < t \leq 40\\
   3.5 \times 10^{-5}                     & \text{if } 40 < t \leq 70\\
   3.5 \times 10^{-6}                     & \text{if } 70 < t \leq 120\\
  \end{cases}
\]
The hyperparameters for the random erasing augmentation (REA)~\cite{Zhong_REA_AAAI_2017} are set to $p = 0.5$; $0.02 < Se < 0.4$; $r_1 = 0.3$; $r_2 = 3.33$. In addition to REA, the BoT model also applies random horizontal flips with probability of $p=0.5$. The label smoothing parameter $\epsilon$ is set to be 0.1, and batch size to 64.

\textbf{Testing.} The BoT model was tested with cosine distance using the features after batch normalization (BN) and k-reciprocal re-ranking~\cite{Zhong_reranking_CVPR_2017}. The batch size is set to 128. We report the Rank1 accuracy and mean Average Precision (mAP) as standard performance evaluation metrics.

 

\begin{table*}[t]
\caption{\label{tab_performance_Market1501} Market1501 performance evaluation.}
\centering\small

\begin{tabular}{l c c c c c c}
\toprule
 & \multicolumn{2}{c}{Trained and tested on} & \multicolumn{2}{c}{Trained on original, } & \multicolumn{2}{c}{Trained and tested on}\\
 
 & \multicolumn{2}{c}{original} & \multicolumn{2}{c}{tested on anonymized} & \multicolumn{2}{c}{anonymized}\\
\cmidrule{2-7}
Models  & mAP & Rank1 & mAP & Rank1 & mAP & Rank1\\
\midrule
PCB          &  72.8 & 87.8 & 71.7 & 87.4 & 72.9 & 88.2\\ 
MLFN         &  71.4 & 87.4 & 70.9 & 86.9 & 71.3 & 87.5\\
HACNN        &  67.4 & 85.5 & 66.5 & 85.7 & 66.6 & 85.2\\ 
Resnet50Mid  &  72.9 & 88.1 & 71.5 & 87.5 & 72.7 & 87.9\\ 
MuDeep       &  44.9 & 70.4 & 43.7 & 70.0 & 44.8 & 69.6\\ 
BoT          &  94.1 & 95.5 &  93.7 &  95.0 &  94.0 & 95.2\\ 

\bottomrule
\end{tabular}
\end{table*}

\begin{table*}[t]
\caption{\label{tab_performance_Duke} DukeMTMC-reID performance evaluation.}
\centering\small

\begin{tabular}{l c c c c c c}
\toprule
 & \multicolumn{2}{c}{Trained and tested on} & \multicolumn{2}{c}{Trained on original, } & \multicolumn{2}{c}{Trained and tested on}\\
 
 & \multicolumn{2}{c}{original} & \multicolumn{2}{c}{tested on anonymized} & \multicolumn{2}{c}{anonymized}\\
\cmidrule{2-7}
Models  & mAP & Rank1 & mAP & Rank1 & mAP & Rank1\\
\midrule

PCB          & 66.3 & 80.3 & 64.4 & 79.3 & 65.4 & 80.2 \\ 
MLFN         & 60.4 & 78.1 & 57.9 & 76.8 & 59.7 & 77.2 \\
HACNN        & 57.4 & 74.0 & 55.5 & 72.2 & 56.7 & 73.1 \\ 
Resnet50Mid  & 62.9 & 80.1 & 60.1 & 78.7 & 61.8 & 80.8\\ 
MuDeep       & 36.0 & 56.6 & 34.4 & 54.4 & 34.8 & 54.7\\ 
BoT          & 88.8 & 90.4 & 87.8 & 89.7 & 88.6 & 90.3\\ 

\bottomrule
\end{tabular}
\end{table*}

\begin{table*}[t]
\caption{\label{tab_performance_VIPeR} VIPeR performance evaluation.}
\centering\small

\begin{tabular}{l c c c c c c}
\toprule
 & \multicolumn{2}{c}{Trained and tested on} & \multicolumn{2}{c}{Trained on original, } & \multicolumn{2}{c}{Trained and tested on}\\
 
 & \multicolumn{2}{c}{original} & \multicolumn{2}{c}{tested on anonymized} & \multicolumn{2}{c}{anonymized}\\
\cmidrule{2-7}
Models  & mAP & Rank1 & mAP & Rank1 & mAP & Rank1\\
\midrule

PCB          & 55.3 & 43.0 & 54.6 & 42.7 & 55.5 & 41.1\\ 
MLFN         & 37.4 & 23.7 & 34.0 & 21.8 & 37.1 & 25.0\\
HACNN        & 25.4 & 14.9 & 25.1 & 14.9 & 25.6 & 15.8\\ 
Resnet50Mid  & 40.9 & 29.1 & 37.9 & 25.3 & 39.5 & 28.5\\ 
MuDeep       & 23.1 & 13.9 & 23.2 & 13.9 & 21.6 & 12.3\\ 
BoT          & 35.9 & 23.4 & 33.0 & 20.3 & 33.8 & 22.5\\ 
\bottomrule
\end{tabular}
\end{table*}

\begin{table*}[t]
\caption{\label{tab_performance_Airport} Airport performance evaluation.}
\centering\small

\begin{tabular}{l c c c c c c}
\toprule
 & \multicolumn{2}{c}{Trained and tested on} & \multicolumn{2}{c}{Trained on original, } & \multicolumn{2}{c}{Trained and tested on}\\
 
 & \multicolumn{2}{c}{original} & \multicolumn{2}{c}{tested on anonymized} & \multicolumn{2}{c}{anonymized}\\
\cmidrule{2-7}
Models  & mAP & Rank1 & mAP & Rank1 & mAP & Rank1\\
\midrule

PCB          & 60.2 & 58.1 & 59.9 & 57.8 & 59.9 & 58.5 \\ 
MLFN         & 63.6 & 61.3 & 63.0 & 61.3 & 62.4 & 60.9 \\
HACNN        & 37.1 & 32.5 & 35.8 & 31.2 & 37.5 & 33.7 \\ 
Resnet50Mid  & 61.6 & 58.5 & 60.6 & 58.0 & 61.1 & 59.2 \\ 
MuDeep       & 21.9 & 20.6 & 20.5 & 18.7 & 19.8 & 16.8 \\ 
BoT          & 62.3 & 60.3 & 56.4 & 51.7 & 61.1 & 59.2 \\ 
\bottomrule
\end{tabular}
\end{table*}

\begin{table*}[t]
\caption{\label{tab_performance_CUHK03} CUHK03 performance evaluation.}
\centering\small

\begin{tabular}{l c c c c c c}
\toprule
 & \multicolumn{2}{c}{Trained and tested on} & \multicolumn{2}{c}{Trained on original, } & \multicolumn{2}{c}{Trained and tested on}\\
 
 & \multicolumn{2}{c}{original} & \multicolumn{2}{c}{tested on anonymized} & \multicolumn{2}{c}{anonymized}\\
\cmidrule{2-7}
Models  & mAP & Rank1 & mAP & Rank1 & mAP & Rank1\\
\midrule

PCB          & 50.4 & 52.9 & 47.5 & 49.5 & 50.4 & 51.1\\ 
MLFN         & 51.8 & 53.4 & 48.6 & 50.7 & 51.7 & 53.6\\
HACNN        & 32.6 & 30.7 & 31.7 & 30.4 & 32.9 & 32.4\\ 
Resnet50Mid  & 52.6 & 54.2 & 49.7 & 53.1 & 51.1 & 52.9\\ 
MuDeep       & 22.5 & 23.4 & 21.2 & 22.4 & 22.2 & 23.3\\ 
BoT          & 58.4 & 61.1 & 67.9 & 64.2 & 56.8 & 59.3 \\ 

\bottomrule
\end{tabular}
\end{table*}

\subsection{Impact of anonymization on performance}
Comparisons between six state-of-the-art models (PCB, MLFN, HACNN, Resnet50Mid, MuDeep, and BoT) on Market-1501, DukeMTMC-reID, VIPeR, Airport and CUHK03 datasets are
shown in Tables~\ref{tab_performance_Market1501}-\ref{tab_performance_CUHK03}. We report results for three cases where we (i) train and test on the original dataset; (ii) train on the original and evaluate on the anonymized dataset; and (iii) train and test on the anonymized dataset. 

We observe that the performance results follow the same pattern for each dataset. The performance is marginally diminished when we train on the original and evaluate all models on the anonymized versions of all datasets. The accuracy is, however, recovered by training on the anonymized versions of all datasets. 

Overall, the difference in performance figures is very small for all datasets. We conduct a paired sample t-test to analyze the performance before and after the anonymization, and to test if the anonymization had a statistically significant effect on the performance. The average mAP across all models and datasets before anonymization is 58.07 and  56.68 after anonymization \text{(-1.39)}, and 57.44 after retraining on the anonymized data \text{(-0.63)}. Anonymization and retraining does have a statistically significant effect on mAP (one-sided paired t-test, $p < 0.01$) but the effect size is small (0.63). Similar conclusions are found for rank-1 accuracy: average rank-1 accuracy before anonymization is 66.69, after is 65.45 (-1.24), and after retraining is 66.25 (-0.44). Again, the difference in mean is significant $(p < 0.05)$, but very small (less than half a percentage point).


\subsection{Effect of different face anonymization techniques}
\begin{figure}[t]
\centering
\includegraphics[scale=1]{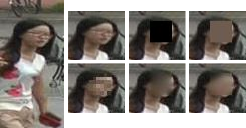}
\caption{Example of different face anonymization techniques on a $64 \times 128$ sample image from the image-based Market1501 dataset. \textbf{Top row}: original face image, blackout on the face detected with the TinyFaces detector, blank; \textbf{Bottom row}: pixelated, inpainted, blurred with a large-kernel Gaussian. }
\label{fig:anon_types}
\end{figure}

This section investigates how different types of face de-identification techniques impact re-ID performance. We focus on the state-of-the-art BoT model and the Market1501 dataset. We consider five anonymization techniques: blur, blank, zero, inpaint, and pixelate. \textit{Blur} uses a Gaussian filter with a radius equal to $1/8$ of the width of the face crop; \textit{zero} replaces the detected face pixels with zero; \textit{blank} replaces detected pixels with the average pixel value; \textit{inpaint} uses Poisson-based image inpainting; and \textit{pixelate} resizes detected region to $(n,n)$ with $n=6$ and then upsamples to the original face crop size using nearest neighbor interpolation.
Figure~\ref{fig:anon_types} shows an example of the above techniques on a sample from the Market1501 dataset.  

\begin{table*}[t]
\caption{\label{tab_face_anon}Performance evaluation of the BoT model on Market 1501 dataset under different face anonymization techniques.}
\centering\small

\begin{tabular}{l c c c c}
\toprule
 & \multicolumn{2}{c}{Trained on original, tested on anonymized} & \multicolumn{2}{c}{Trained and tested on anonymized}\\
\cmidrule{2-5}
Face anon. tech.  & mAP & Rank1 & mAP & Rank1 \\
\midrule
Blur      &  93.7 &  95.0 &  94.0 & 95.2 \\ 
Zero      &  93.4 &  94.6 &  93.8 &  94.5 \\
Blank     &  93.7 & 95.0 & 93.8 & 95.7 \\ 
Pixelate  &  94.0 & 95.3 & 94.0 & 95.4 \\ 
Inpaint   & 93.6 & 95.0 & 93.6 & 95.0 \\ 
\bottomrule
\label{tab:anon_techniques}
\end{tabular}
\end{table*}

The results of this experiment are summarized in Table~\ref{tab:anon_techniques}. We observe that most methods provide comparable results, with \textit{blur} and \textit{pixelate} being slightly superior. The results are somewhat better when trained and tested on the anonymized versions of the Market1501 dataset. Zeroing the faces (replacing all  face pixels with zero) is only 0.25\% on average below \textit{blur}. Therefore, the complete zeroing of the face can be used instead of blurring in cases when extra security is needed.

\section{Discussion}
\textbf{Resistance to deblurring and other attacks.} Existing SoA for deblurring faces (e.g.~\cite{deblur_1_Chrysos_IJCV_2019,deblur_2_Chrysos_CVPRW_2017,deblur_3_Shen_ICCV_2019}) are focused on either removing a type of structured face blurring that occurs due to motion (e.g.~\cite{deblur_1_Chrysos_IJCV_2019,deblur_3_Shen_ICCV_2019}) or relatively minor blurring that can occur due to camera defocus and similar effects (e.g.~\cite{deblur_2_Chrysos_CVPRW_2017}). The structure present in motion blur offers the opportunity to achieve very good results in removing this type of distortion, but these techniques do not work on strong (large kernel, high variance) Gaussian blurs, which simply remove all high-frequency components in the Fourier spectrum. Techniques such as~\cite{deblur_2_Chrysos_CVPRW_2017} also cannot recover faces that have been blurred using strong Gaussian blurs, since the information about the detail in the face is effectively completely removed. 

Image super resolution techniques (e.g.~\cite{deblur_4_Sun_IEEE_2019}) are effective for generating realistic high-resolution content from low-resolution imagery. These techniques, however, effectively only hallucinate a plausible realistic high-resolution version of the low-resolution contents, and therefore cannot be used to reveal the true anonymized face.

Gross et al.~\cite{18_Gross_CVPRW_2006} argue that pixelation and blurring are insufficient to hide identity, and that more sophisticated techniques are needed. They show that images that are pixelated and blurred can still be matched/classified correctly against a dataset of similarly distorted images. However, the dataset used is relatively small and simple (275 subjects), and the matching techniques are outdated (PCA based), which indicates that the subjects were being matched based on relatively coarse-grained characteristics (e.g. hair color or similar surrounds). The paper also indicates that very strong blurring (as we have done) strongly damages recognition ability.

Note that Google uses similar image blurring technology in their Google Street View system to blur faces and license plates and that this technology has been deployed since 2009 without any major incidents in terms of deblurring the images.

\section{Conclusion}

In this study, we empirically observed that the effect of blurring faces on the person re-identification performance is surprisingly small. We also find that the relative performance of different state-of-the-art methods is preserved after anonymization, meaning that new approaches can be safely compared using anonymized data. Going forward, we believe this provides important guidance for future person re-identification research with anonymized datasets. One possible line of future work will be to investigate privacy-preserving face de-identification methods such as those from the GAN-based family of algorithms.

\section*{Acknowledgment}

This publication has emanated from research conducted with the financial support of Science Foundation Ireland (SFI) under grant number SFI/15/SIRG/3283 and SFI/12/RC/2289\_P2. We thank Bhartendu Sharma for his assistance in running several experiments. We would also like to thank Dublin Airport Authority for their contributions to this work.



\end{document}